\documentclass[pdflatex,sn-mathphys-num]{sn-jnl}

\usepackage{graphicx}%
\usepackage{multirow}%
\usepackage{amsmath,amssymb,amsfonts}%
\usepackage{amsthm}%
\usepackage{mathrsfs}%
\usepackage[title]{appendix}%
\usepackage{xcolor}%
\usepackage{textcomp}%
\usepackage{manyfoot}%
\usepackage{booktabs}%
\usepackage{algorithm}%
\usepackage{algorithmicx}%
\usepackage{algpseudocode}%
\usepackage{listings}%
\usepackage{acro}

\DeclareAcronym{3d}{
    short   =   3D,
    long    =   three-dimensional,
}

\DeclareAcronym{3dmm}{
    short   =   3DMM,
    long    =   3D Morphable Model,
}

\DeclareAcronym{ai}{
    short   =   AI,
    long    =   Artificial Intelligence,
}

\DeclareAcronym{cnn}{
    short   =   CNN,
    long    =   Convolutional Neural Network,
}

\DeclareAcronym{cv}{
    short   =   CV,
    long    =   Computer Vision,
}

\DeclareAcronym{dnn}{
    short   =   DNN,
    long    =   Deep Neural Network,
}

\DeclareAcronym{ema}{
    short   =   EMA,
    long    =   Exponential Moving Average,
}

\DeclareAcronym{hpe}{
    short   =   HPE,
    long    =   Head Pose Estimation,
}

\DeclareAcronym{mae}{
    short   =   MAE,
    long    =   Mean Absolute Error,
}

\DeclareAcronym{mse}{
    short   =   MSE,
    long    =   Mean Squared Error,
}

\DeclareAcronym{mlp}{
    short   =   MLP,
    long    =   Multilayer Perceptron,
}

\DeclareAcronym{nme}{
    short   =   NME,
    long    =   Normalized Mean Error,
}

\DeclareAcronym{pnp}{
    short   =   PnP,
    long    =   Perspective-n-Point,
}

\DeclareAcronym{rnn}{
    short   =   RNN,
    long    =   Recurrent Neural Network,
}

\DeclareAcronym{s4}{
    short   =   S4,
    long    =   Structured State Space Sequence Model,
}

\DeclareAcronym{ssl}{
    short   =   SSL,
    long    =   Semi-Supervised Learning,
}

\DeclareAcronym{ssr}{
    short   =   SSR,
    long    =   Semi-Supervised Regression,
}

\DeclareAcronym{ssrr}{
    short   =   SSRR,
    long    =   Semi-Supervised Rotation Regression,
}

\DeclareAcronym{ssm}{
    short   =   SSM,
    long    =   State Space Model,
}

\DeclareAcronym{vim}{
    short   =   Vim,
    long    =   Vision Mamba,
}

\DeclareAcronym{vit}{
    short   =   ViT,
    long    =   Vision Transformer,
}

\DeclareAcronym{gan}{short=GAN, long=Generative Adversarial Network}

\DeclareAcronym{vae}{short=VAE, long=Variational Autoencoder}

\theoremstyle{thmstyleone}%
\theoremstyle{thmstyletwo}%

\theoremstyle{thmstylethree}%

\begin{document}

\title[Article Title]{Bengal-HP\_RU: A Dataset of Bengal People For Head Pose Estimation}


\author[1]{\fnm{Md. Ahanaf Arif} \sur{Khan}}\email{s1910676110@ru.ac.bd}
\author[1]{\fnm{Md. Tawhidur} \sur{Rahman}}\email{s2310776126@ru.ac.bd}

\author*[1]{\fnm{Sangeeta} \sur{Biswas}}\email{sangeeta.cse@ru.ac.bd}
\author[1]{\fnm{Md. Iqbal Aziz} \sur{Khan}}\email{iqbal\_aziz\_khan@ru.ac.bd}

\author[1]{\fnm{Subrata} \sur{Pramanik}}\email{sprmnk@ru.ac.bd}

\author[1]{\fnm{Sanjoy Kumar} \sur{Chakravarty}}\email{sanjoy.cse@ru.ac.bd}

\author[1]{\fnm{Bimal Kumar} \sur{Pramanik}}\email{bkp@ru.ac.bd}

\affil[1]{\orgdiv{Department of Computer Science and Engineering, Faculty of Engineering}, \orgname{University of Rajshahi}, \orgaddress{\city{Rajshahi}, \postcode{6205}, \country{Banglasesh}}}


\abstract{Existing head pose datasets predominantly feature subjects of Western or East Asian origin, leaving South Asian populations, particularly Bengali individuals, largely underrepresented. We introduce Bengal-HP\_RU, the first publicly available head pose dataset centred on Bengali subjects, comprising 12,894 labelled head images annotated with continuous yaw, pitch, and roll values. Images were collected from Wikimedia Commons under free licences and processed through an automated pipeline followed by manual label correction. The dataset is partitioned by Wikimedia uploader identity to prevent data contamination, yielding 10,494 training and 2,400 test images across 296 unique uploaders. Bengal-HP\_RU exhibits substantial diversity in subject age, gender, occlusion, illumination, and background, reflecting realistic in-the-wild conditions. The dataset is publicly available at \url{https://doi.org/10.17632/xbw9kr37jb.2}.}

\keywords{Head Pose Estimation, Dataset, Bengali, South Asian, In-the-Wild}



\maketitle

\section{Introduction}\label{sec1}

\begin{figure}[!h]
    \centering
    \includegraphics[width=0.9\textwidth]{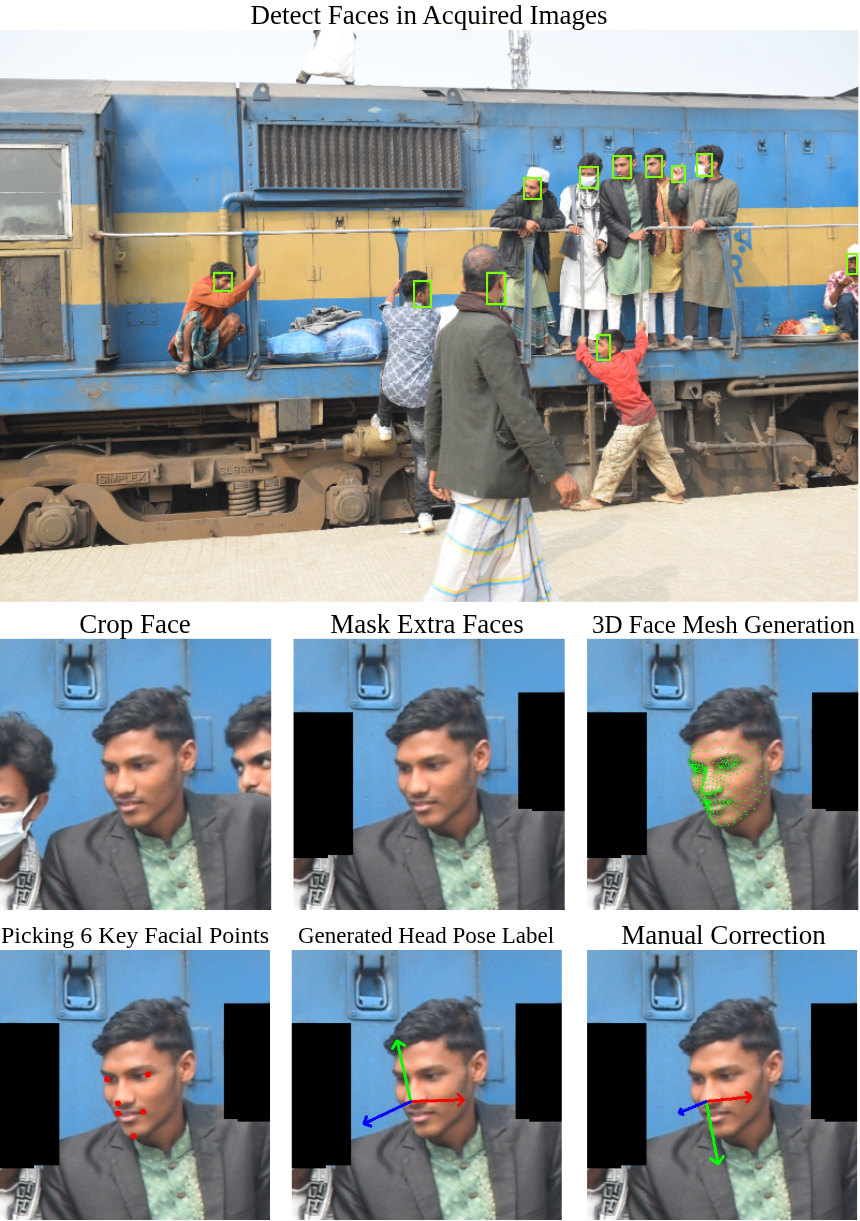}
    \caption{Overview of the Bengal-HP\_RU dataset creation pipeline: face detection and cropping, masking of extra faces, 3D face mesh generation, 6-point pose estimation via \texttt{solvePnP}, and final manual label correction.}
    \label{fig:label-pipeline}
\end{figure}

Accurate head pose estimation depends critically on the availability of large-scale, diverse, and representative training data. While several head pose datasets exist in the literature~\cite{DB_BIWI_Fanelli2012,DB_300W_AFLW2000_Zhu2016_CVPR,DB_200wlpa_Hsu2020_Q1,DB_DAD3D_Martyniuk2022_Q1}, they predominantly feature subjects of Western or East Asian origin, with little to no representation of South Asian populations, particularly Bengali individuals. Facial geometry, skin tone distribution, and cultural appearance attributes (such as attire and accessories) vary across ethnic groups, and models trained on non-representative datasets often generalise poorly to underrepresented demographics. To address this gap, we introduce {Bengal-HP\_RU}, a large-scale head pose dataset curated specifically to capture the demographic and contextual diversity of Bengali individuals. Bengal-HP\_RU is, to the best of our knowledge, the first publicly available head pose dataset centred on Bengali subjects. It offers in-the-wild image conditions, rich diversity in age, gender, occlusion patterns, and illumination, making it a valuable resource for training and evaluating robust head pose estimation models in real-world scenarios. Figure~\ref{fig:label-pipeline} illustrates the complete dataset creation pipeline of Bengal-HP\_RU, from face detection in the acquired image through to manual correction of the final label.

\section{Image Acquisition}

All images in Bengal-HP\_RU were collected from Wikimedia Commons, a repository of freely licensed media content. We conducted targeted searches using Bengali-relevant keywords and categories and retrieved a total of 3,591 images depicting individuals of Bengali origin in unconstrained, in-the-wild settings. For each collected image, we also retrieved associated metadata, including the original filename, title, image URL, Wikimedia page link, uploader information, upload and original capture dates, image description, artist attribution, licensing details, and categorical tags. The in-the-wild nature of the source ensures that the dataset reflects realistic imaging conditions, encompassing natural variation in pose, background clutter, partial occlusion, and ambient lighting, conditions that are frequently encountered in practical deployment scenarios but often underrepresented in laboratory-collected datasets.

\section{Image Processing}

Following image acquisition, each image was passed through an automated processing pipeline to isolate individual face crops suitable for head pose annotation. Face detection was performed using the YOLOv8-Face medium model \cite{yusepp_yolov8_face}, which provides accurate bounding box predictions across a wide range of face sizes, orientations, and occlusion levels. Detected faces were then cropped with a standardised margin. To prevent label ambiguity and training noise, images containing multiple detected faces were masked such that non-target faces were excluded from the visible region. This ensured that each processed sample corresponded to a single, unambiguous subject.

\section{Train/Test Split}

To prevent data contamination between the training and evaluation subsets, the dataset was partitioned based on the Wikimedia Commons uploader identifier. Since individual uploaders often contribute multiple images of the same subject or from the same event, a subject-aware or session-aware split based on uploader identity ensures that images originating from the same source do not appear in both the training and test sets simultaneously. The split was performed using a greedy algorithm that first sorts uploaders by their total image count in descending order, then assigns uploaders to the training set until the 80\% image target is reached, with all remaining uploaders assigned to the test set. This approach offers two complementary advantages over a naive random split: it prevents images of the same subject or event from leaking across the train/test boundary, and by concentrating high-volume uploaders in training, it reserves a broad spread of unseen uploaders for evaluation, better reflecting generalisation to new, previously unencountered sources. The resulting split comprises 2,876 training source images and 715 test source images, yielding 10,494 and 2,400 cropped head images respectively, totalling 12,894 head images across 296 unique uploaders, as summarised in Table~\ref{tab:train-test-split}.

\begin{table}[!h]
    \centering
    \caption{Train/test split statistics of the Bengal-HP\_RU dataset.}
    \label{tab:train-test-split}
    \begin{tabular}{llllrlllrlllr}
        \hline
        &   & &&\textbf{Train}  & & && \textbf{Test}  & & && \textbf{Total} \\
        \hline
        \textbf{Uploaders}    &   & &&21      & & && 275    & & && 296    \\
        \textbf{Source Images} &   & &&2,876   & & && 715    & & && 3,591  \\
        \textbf{Head Images}  &   & &&10,494  & & && 2,400  & & && 12,894 \\
        \hline
    \end{tabular}
\end{table}

\section{Head Pose Label Generation}
We generated head pose labels through a two-stage annotation process combining automated geometric estimation with manual correction. Each image was first processed through a pipeline to extract face landmarks and compute pitch, yaw, and roll angles. These automatically derived angles were then used as a reference during a manual review stage, where every image was inspected and labeled individually. 

\subsection*{Stage 1: Automated Pose Estimation}
We built an automated pipeline using MediaPipe and OpenCV to process each image and estimate the head pose from facial geometry. The pipeline runs sequentially through landmark detection, key point extraction, 3D pose solving, and angle computation before saving the results per image.

\begin{itemize}
    \item \textbf{Step 1: Detecting Face Landmarks}
    
    First, we load each image using OpenCV. We then use MediaPipe's~\cite{Mediapipe_Lugaresi2019_arXiv} FaceMesh model on the image and generate 478 normalized landmarks. These landmarks are then converted into real pixel coordinates by multiplying $x$ by the image width and $y$ by the image height, with the $z$ depth also scaled by width.
    
    \item \textbf{Step 2: Picking Six Key Facial Points for Pose Estimation}
    
    From the 478 landmarks, six specific anatomical points are extracted to be used in the pose estimation step. These points are: the nose tip, chin, left eye corner, right eye corner, left mouth corner, and right mouth corner. These six points serve as the 2D image points that anchor the geometry calculation in the next step.
    
    \item \textbf{Step 3: Running \texttt{solvePnP}}
    
    We define a canonical 3D face model as a fixed set of idealized 3D coordinates for the same 6 facial landmarks, representing an average human face in 3D space. We also construct a camera matrix assuming a simple pinhole camera, with the focal length approximated as the image width and the principal point placed at the image center. We then apply OpenCV's \texttt{solvePnP} to solve the \ac{pnp} problem, i.e., given the 6 known 3D model points and their corresponding 2D pixel locations, it estimates the rotation and translation that best explain the observed face configuration. The output is a rotation vector and a translation vector.
    
    \item \textbf{Step 4: Converting to Euler Angles}
    
    We convert the rotation vector from \texttt{solvePnP} into a $3 \times 3$ rotation matrix using \texttt{cv2.Rodrigues}, which is then decomposed into pitch, yaw, and roll using OpenCV's \texttt{RQDecomp3x3}. We normalize all angles to the range $[-180^\circ, 180^\circ]$ and apply a correction for flipped solutions, a known ambiguity in \texttt{solvePnP}, by mirroring pitch or roll back to their stable equivalent when either exceeds $\pm 90^\circ$. The face mesh (478 landmarks),  68 most informative landmarks, selected from the full set of 478, are saved alongside the face bounding box and the computed Euler angles in a per-image \texttt{.npz} file for use in downstream tasks.
\end{itemize}

\subsection*{Stage 2: Manual Label Correction}
\begin{figure}[!h]
    \centering
    \includegraphics[width=\textwidth]{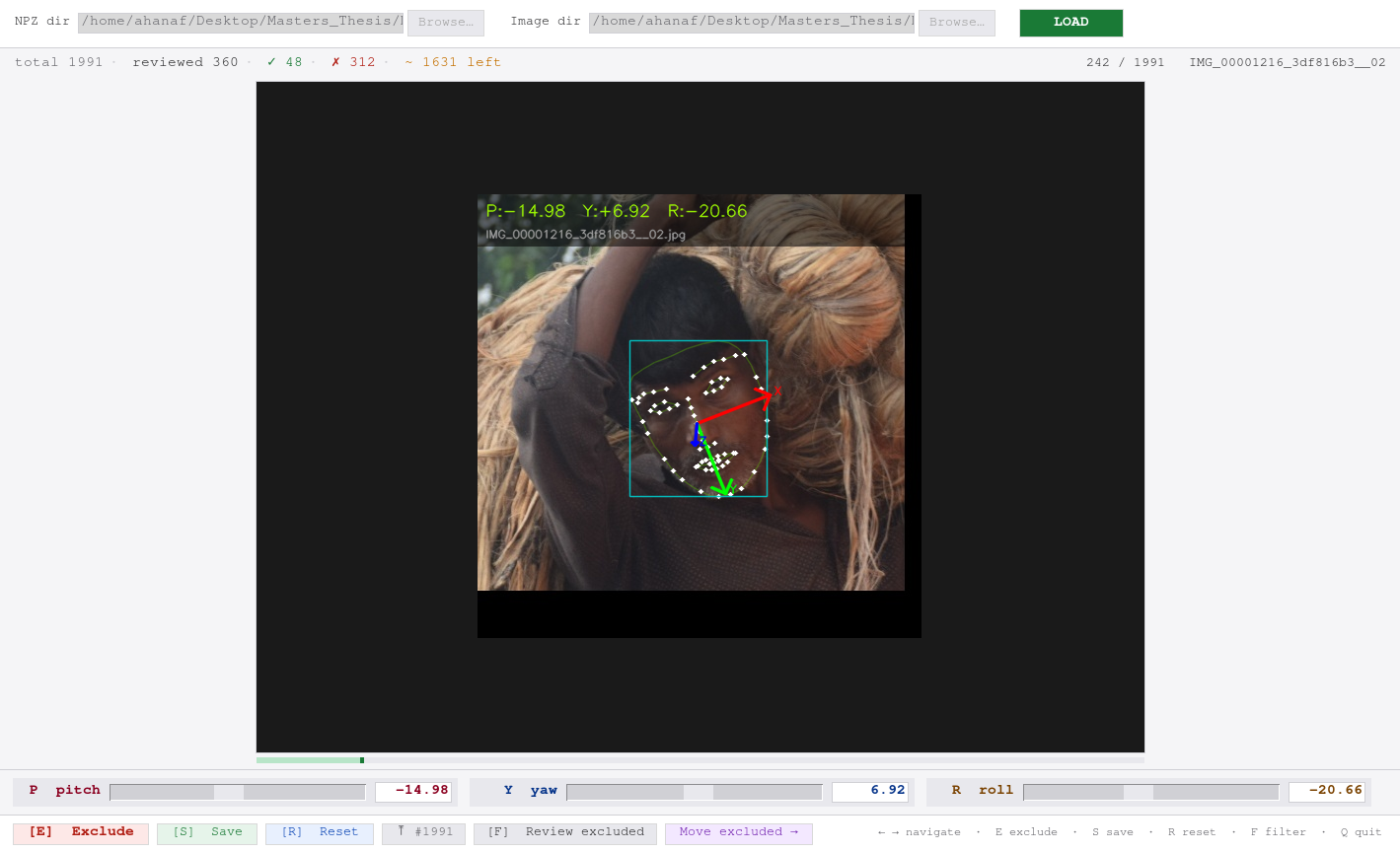}
    \caption{Our custom head pose annotation tool. For each image, the tool overlays the 68 facial landmarks and a 3D pose axis (red, green, and blue arrows) on the detected face, along with the computed pitch (P), yaw (Y), and roll (R) values. Annotators could adjust the angles using the sliders at the bottom and save, reset, or exclude each image individually.}
    \label{fig:annot-tool}
\end{figure}

Since \texttt{solvePnP} relies on an idealized 3D face model and a rough camera approximation, the automatically computed angles can be unreliable on borderline or unusual head orientations. To address this, we developed a custom head pose annotation tool (Figure~\ref{fig:annot-tool}) that displayed each image alongside its automatically computed angles and allowed annotators to adjust the label manually. The 68 facial landmarks overlaid on each image played a key role during this process. They provided a precise visual reference of the face geometry, making it easier to judge the true head orientation even in ambiguous or extreme poses. The entire reannotation was carried out by a single annotator to ensure consistency across the dataset. Additionally, images that were flagged during the automated stage for having no detected face, false detections, or visibly poor quality were reviewed and excluded from the dataset entirely to ensure annotation reliability.

\section{Dataset Characteristics}

\begin{figure}[h]
    \centering
    \includegraphics[width=\linewidth]{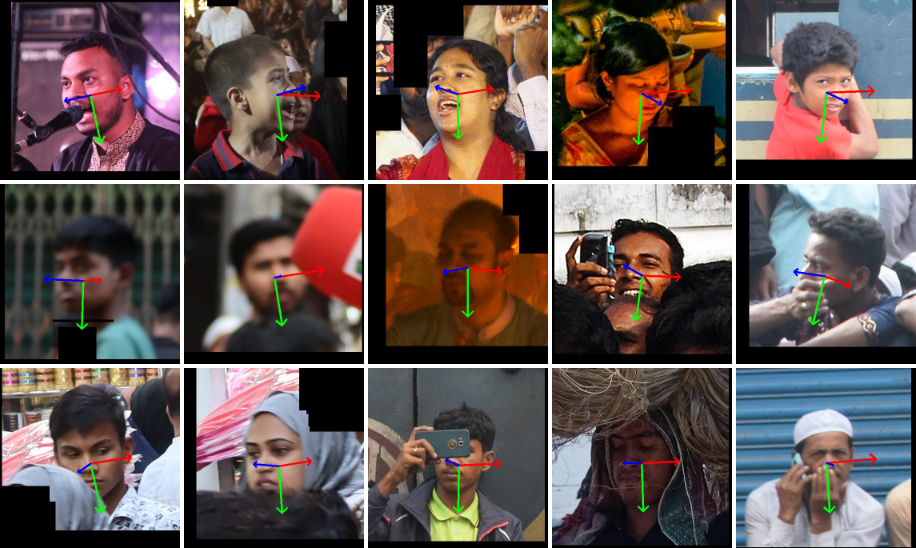}
    \caption{Representative samples from the Bengal-HP\_RU dataset demonstrating the diversity in subject demographics, backgrounds, image quality, occlusion patterns, and illumination conditions.}
    \label{fig:bengalhp_samples}
\end{figure}

The resulting Bengal-HP\_RU dataset comprises a total of 12,894 labelled head images annotated with continuous yaw, pitch, and roll values. The dataset exhibits substantial diversity along several axes, including subject age, gender, image resolution, background and lighting conditions, and degree of facial occlusion. Representative samples illustrating this diversity are shown in Figure~\ref{fig:bengalhp_samples}.

\section{Dataset Structure}

The dataset is organised into two subsets, train and test, each containing a directory of cropped head images and a corresponding directory of per-image \texttt{.npz} files. The \texttt{.npz} files store the 68 facial landmarks, face bounding box, and computed Euler angles for each image. The metadata of the source images for each subset is provided in \texttt{train.csv} and \texttt{test.csv}, respectively. The full directory structure is as follows:

\begin{verbatim}
Bengal-HP_RU/
|-- train.csv
|-- test.csv
|-- demo.py                  # Example script for loading NPZ annotations 
|                               and visualizing them on an image
|-- train/
|   |-- images_train/        # 10,494 cropped head images
|   `-- npz_train/           # 10,494 per-image .npz files
`-- test/
    |-- images_test/         # 2,400 cropped head images
    `-- npz_test/            # 2,400 per-image .npz files
\end{verbatim}

\section{Limitations}
The Bengal-HP\_RU dataset has several limitations worth noting. The facial landmarks stored in the \texttt{.npz} files are generated automatically by MediaPipe and are not manually corrected, and may therefore be inaccurate in challenging poses or occlusion conditions. Finally, manual annotation was performed by a single annotator, which, while ensuring label consistency, introduces the risk of individual bias in the final pose labels.

\section{Ethical Considerations and Data Integrity}

All images in Bengal-HP\_RU were sourced exclusively from Wikimedia Commons and are covered by free and open licences permitting reuse. No images were collected through direct subject recruitment, and no personally identifiable information beyond what is publicly available on Wikimedia Commons is retained or distributed. The dataset is intended solely for non-commercial computer vision research.

\section{Dataset Availability}

The Bengal-HP\_RU dataset is publicly available on Mendeley Data at \url{https://doi.org/10.17632/xbw9kr37jb.2}.

\backmatter

\bigskip    
\bigskip    



\bibliography{references}

\end{document}